\documentclass{article}
\usepackage{spconf,amsmath,graphicx}

\usepackage{comment}
\usepackage{graphicx}
\usepackage{booktabs}
\usepackage{floatrow}
\usepackage{subcaption}  
\usepackage{tikz}
\usetikzlibrary{positioning}
\usetikzlibrary{fit,calc}  
\usepackage{multirow}
\usepackage{colortbl}
\usepackage[T1]{fontenc}
\usepackage[font=small,labelfont=bf,tableposition=top]{caption}
\usepackage{algorithm}  
\usepackage{algpseudocode}  
\usepackage{sidecap} % side caption for table

\DeclareCaptionLabelFormat{andtable}{#1~#2  \&  \tablename~\thetable}

\definecolor{cvprblue}{rgb}{0.21,0.49,0.74}
\usepackage[pagebackref,breaklinks,colorlinks,citecolor=cvprblue]{hyperref}

\title{Efficient Text-to-Image Generation: An Adaptive Step Schedule Controller for Diffusion Models}
\address{}
\name{Kuluhan Binici\textsuperscript{\rm 1, \rm 2},
    Cihan Acar\textsuperscript{\rm 3},
    Shivam Aggarwal\textsuperscript{\rm 2},
    Siying Liu\textsuperscript{\rm 3},
    Tulika Mitra\textsuperscript{\rm 2}}
\address{\textsuperscript{\rm 1}SAP,
\textsuperscript{\rm 2}National University of Singapore,
\textsuperscript{\rm 3}Institute for Infocomm Research (I2R), A*STAR, Singapore}
\begin{document}
%\ninept
%
\maketitle
\begin{abstract}
 Text-to-image diffusion models often use a fixed number of denoising steps, balancing time costs and image quality. However, the optimal number of steps depends on the complexity of the input text prompt. We propose an adaptive diffusion controller that dynamically adjusts the number of steps to generate high-quality images efficiently, without additional model training. By leveraging a mixture of step schedules with varying step sizes and evaluating the error term discrepancy at each timestep, our method transitions between schedules to optimize performance. Experiments on COCO and DiffusionDB show that our approach reduces inference time while maintaining visual fidelity, offering a more efficient alternative for text-to-image diffusion models.
\end{abstract}
\vspace{-5pt}
\section{Introduction}
\label{sec:intro}
\begin{figure}[!ht]
  \centering
  \begin{subfigure}{1\textwidth}
  \centering
        \includegraphics[width=.48\linewidth]{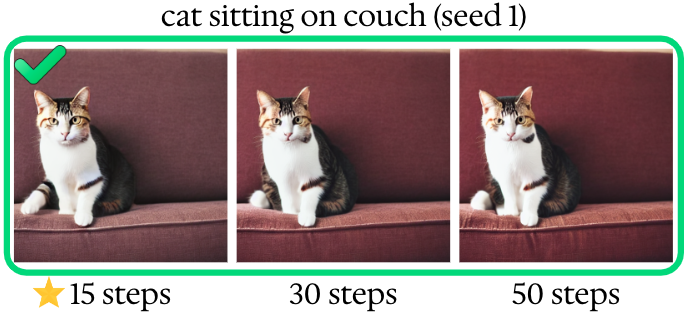}
        \includegraphics[width=.48\linewidth]{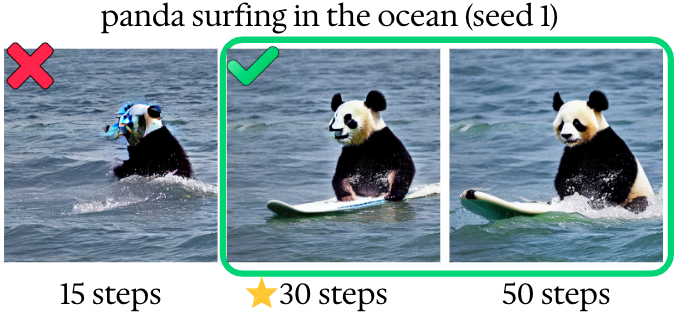}
   \end{subfigure}
   \begin{subfigure}{1\textwidth}
   \centering
   \includegraphics[width=.48\linewidth]{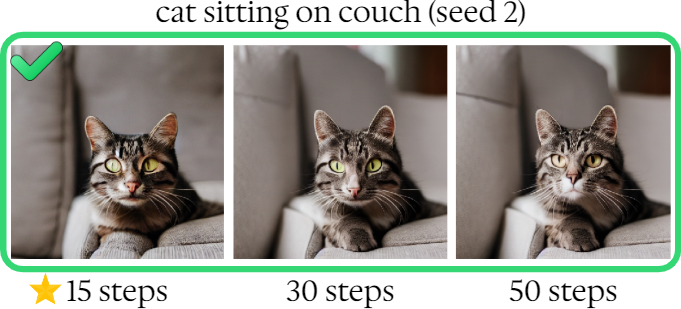}
   \includegraphics[width=.48\linewidth]{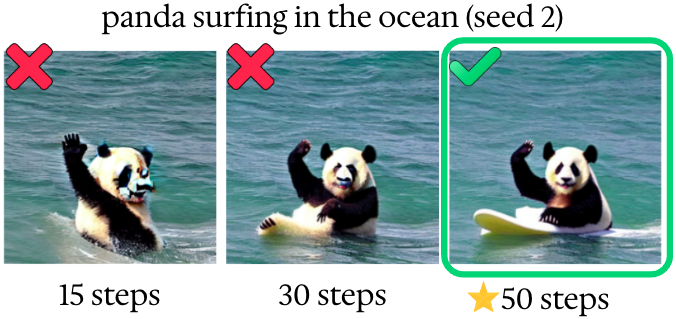}
   \end{subfigure}
   \vspace{-15pt}
   \caption{Effect of text prompt and seed combinations on diffusion steps for generating visually sufficient images. Step counts shown are examples; optimal steps may vary.}
   \label{fig:onecol}
\end{figure}

Diffusion models have showcased an impressive ability to create high-diversity and high-quality images from textual descriptions \cite{nichol2021glide,ramesh2022hierarchical}. 
%\cite{nichol2021glide,ramesh2022hierarchical,ramesh2021zero,rombach2022high,saharia2022photorealistic}. 
One of the primary challenges with diffusion models is the high computational time incurred during the inference process. 
Knowledge distillation and neural architecture search techniques have been implemented to decrease the required number of time steps in the generation process \cite{luhman2021knowledge,kim2023architectural}. 
However, these require supplementary training of models.
In addition, these methods typically use a fixed number of diffusion steps without considering the specific needs or complexity of the task. This one-size-fits-all approach to the denoising schedule overlooks the nuanced dynamics of the diffusion process, which can lead to either redundant computation or potentially immature results.
 As illustrated in Figure \ref{fig:onecol}, the optimal number of steps is affected by the initialization of the noisy image, i.e. the random seed, and the text prompt. 
Notably, the dependence of the optimal time step schedule for the specific generation task has also been acknowledged by previous works \cite{li2023autodiffusion}.
In this paper, we address this inefficiency by proposing an adaptive controller designed to allocate a minimal number of steps required to achieve visually satisfactory results for each text prompt. This allows better visual quality retention while improving time efficiency, rather than simply reducing the fixed schedule for all images. We named our approach as \textit{Adaptive Diffusion Step Controller} (ADSC). ADSC does not require additional model training and functions by monitoring the convergence of the generation process during runtime to adjust the schedule accordingly. We postulate that the cosine distance between the conditional and unconditional update directions can serve as a heuristic for estimating convergence. When the cosine distance is large, indicating that the content described by the text prompt has not been generated in the denoised image yet, our scheduler opts for taking smaller step sizes to refine the generative process with finer granularity. Conversely, when the distance narrows, we infer that such content information is already contained in the image; thus taking larger step sizes are employed, expediting the overall process without compromising the image quality.
Our observations reveal a discernible pattern where the cosine distance between update directions is initially substantial, reflecting the high entropy in the early stages  that the diffusion process has not converged and the denoised image has not aligned with the text prompt. This distance typically diminishes as the process unfolds, permitting the scheduler to incrementally increase the step size. 

To validate the efficacy of our proposed method, we conducted extensive experiments on two well-established datasets: COCO \cite{lin2014microsoft} and DiffusionDB \cite{wang2022diffusiondb}. 
For evaluation metric, we utilize a variant of the CLIP score \cite{hessel2021clipscore}; namely CLIP-I \cite{ruiz2023dreambooth}, and also the DINO metric \cite{caron2021emerging,ruiz2023dreambooth}. These metrics mainly compare the feature-space projections of the generated and reference images. The experimental results reveal that our adaptive approach can effectively reduce the average number of timesteps required to generate images from a given sequence of text prompts, in some cases by more than 50\%, all while maintaining the visual quality. 
\vspace{-5pt}
\section{Related Work}
\label{sec:formatting}
\vspace{-5pt}
To direct the generative process towards predetermined attributes or categories, external classifiers were incorporated to diffusion models but this requires the training and integration of distinct classifier models.
Classifier-free diffusion guidance \cite{ho2022classifier} effectively tackles this by embedding the guidance mechanism within the diffusion model itself, eliminating the need for an external classifier. 
%This adaptation involves altering the diffusion model's training regime to intrinsically learn the generation of data conditioned on specific attributes or categories. 
Subsequent studies have enhanced the image generation process more precisely by guiding the internal representations within the diffusion models to control attributes like shape, location, and appearance of objects \cite{epstein2024diffusion} and also by incorporating additional inputs such segmentation maps, keypoints and bounding boxes to introduce spatial conditioning controls \cite{li2023gligen,chen2024training}.     
Due to the iterative nature of the denoising process, diffusion models face a significant issue in that they require a considerable amount of computational time during the inference process, as multiple forward passes are needed for each image generation. 
%In contrast, Generative Adversarial Networks (GANs)  \cite{goodfellow2014generative} are capable of producing a series of images through a single evaluation, showcasing a significant efficiency advantage.
To tackle this challenge, efficient solvers such as DDIM \cite{song2020denoising}, PNDM \cite{liu2022pseudo} and LMS \cite{karras2022elucidating} has been introduced to accelerate the sampling process.
In addition, architectural compression \cite{kim2023architectural}, quantization \cite{li2023q}, and knowledge distillation techniques have been implemented to decrease the required number of time steps in the generation process \cite{luhman2021knowledge,meng2023distillation}.
Recently, Denoising Diffusion Step-aware Models (DDSMs) \cite{yang2023denoising} have been developed, utilizing neural networks of varying sizes tailored to the specific demands of different stages of the generation process. 
%This approach allows for a more efficient and adaptive generation by aligning the computational resources with the complexity of each step.
The main drawback of these techniques is the need for supplementary training of models specifically for distillation, as well as the requirement to undertake searches for the most effective model architectures \cite{li2023autodiffusion}.
Furthermore, these methods employ a fixed number of diffusion steps, overlooking the need to adapt to the distinct challenges posed by different prompts. 
\vspace{-7pt}
\section{Adaptive Diffusion Step Controller}
\vspace{-5pt}
Our Adaptive Diffusion Step Controller (ADSC) encompasses two key technical components. First, we establish a diffusion step schedule control space within which our ADSC operates. This space is comprised of various schedules with distinct step lengths, enabling the controller to determine an optimal configuration by intelligently blending these schedules. Importantly, the step schedule control space allows each generation task to be completed in a varying number of steps, depending on the requirements of the generation task. Secondly, we introduce a heuristic criterion that guides the ADSC controller in crafting schedules tailored to each text prompt. This criterion strives to estimate the convergence of the diffusion process by monitoring the cosine distances between the conditional and unconditional noise estimations of the diffusion model.
\vspace{-5pt}
\subsection{Heterogenous schedule control space} 
\vspace{-5pt}
\begin{figure}[!ht]
  \centering
    \vspace{-5pt}
\includegraphics[width=0.9\linewidth]{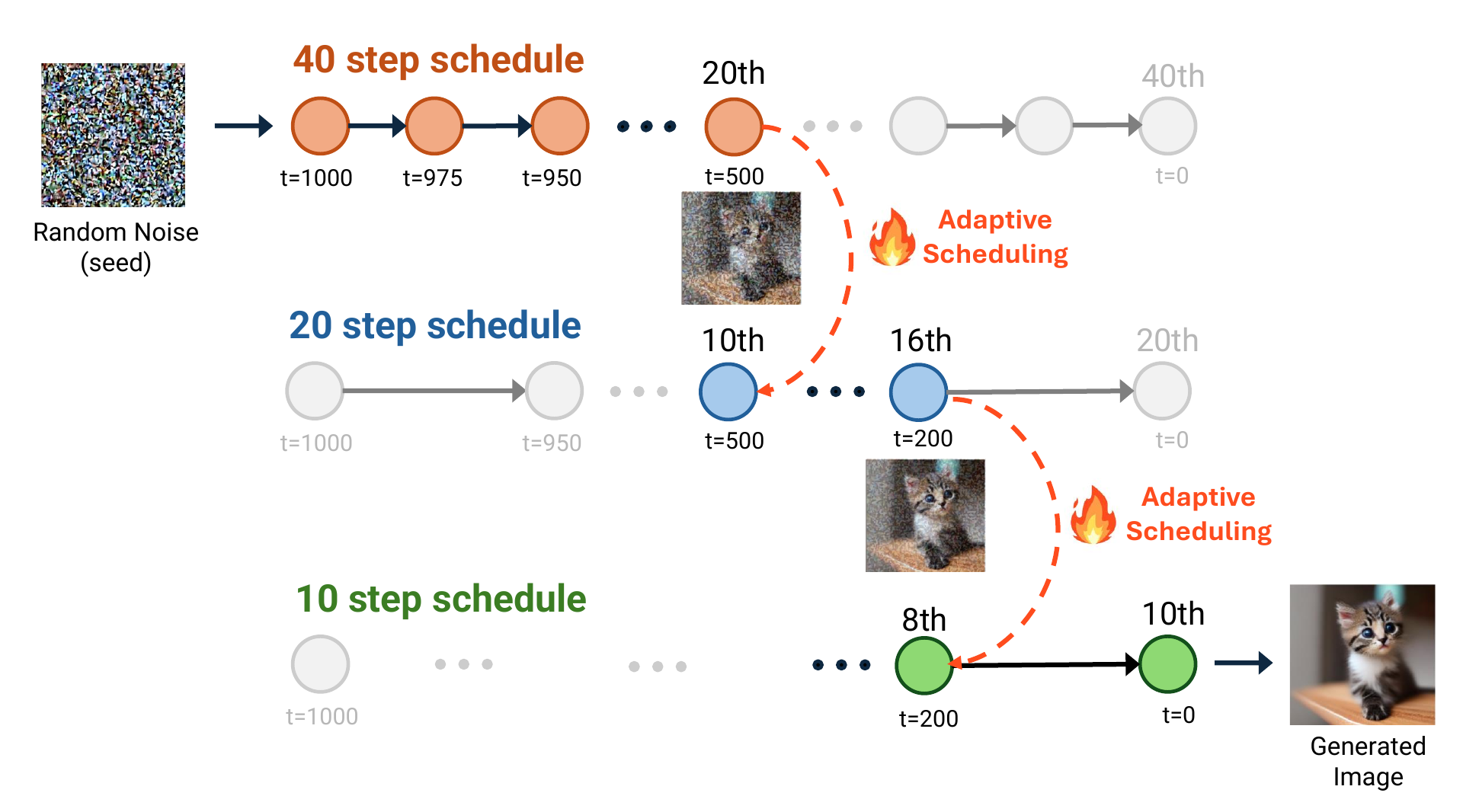} 
           \caption{The denoising process begins with the longest schedule (e.g., 40 steps in the example) and transitions to shorter schedules (20 followed by 10) upon detecting convergence.}
   \label{fig:framework}
   \vspace{-10pt}
\end{figure}
The control space consists of K schedules, each with a varying number of steps, denoted as $S^{k} = \{t^{k}_{1}, t^{k}_{2}, ... t^{k}_{N^k}\}$, where $|S^{k}| = N^k$ represents the number of steps in schedule $k\in[1,K]$. The difference in time units between the steps in each schedule is distinct, resulting in diverse step sizes across the schedules. The controller initiates the execution by following the schedule with the most number of steps. It transitions to the corresponding step in the subsequent schedule whenever convergence is flagged. Each schedule $S^{k}$ has fewer steps and consequently larger step sizes compared to the previous one :
\begin{align}
    N^{k} > &N^{k+1} \\
    t^{k+1}_{i+1} - t^{k+1}_{i} > t^{k}_{j+1} - t^{k}_{j} > 0; & \quad \forall i \in [1,N^{k+1}), \forall j \in [1,N^{k})
\end{align}
This is motivated by the presumption that as the diffusion process converges and a major portion of the content information has been generated, the estimation error incurred by taking larger steps could have less impact on the final image quality.Moreover, the control space can be represented by $C = \{S^{1}, S^{2}, ..., S^{K}\}$. When the transition criterion is met during the execution of the $t^{k}_{i}$ timestep of a certain schedule $S^{k}$, the controller transitions to the subsequent schedule and resumes diffusion starting from the timestep $t^{k+1}_{j}$ that is closest to $t^{k}_{i}$, i.e. 
\begin{equation}
j = \arg\min\limits_{j'} |t^{k+1}_{j'} - t^{k}_{i}|
\end{equation}
%This strategy ensures minimal differences between steps during transitions between schedules. 
Finally, the set of all possible mixtures of schedules that can be generated by the controller through its exploration on the control space, can be formulated as:
\begin{comment}
 \begin{equation}
M = \bigcup_{k=1}^{K}\{t^{k}_{st}, t^{k}_{st+1}, ..., t^{k}_{end}\} \quad \text{where} \quad t^{k}_{1} \leq t_{st} \leq t_{end} \leq t^{k}_{N^{k}}
\end{equation}

\begin{equation}
M = \bigcup_{k=1}^{K}\{t^{k}_{st}, t^{k}_{st+1}, ..., t^{k}_{end}\} \quad ; \quad t^{k}_{1} \leq t_{st} \leq t_{end} \leq t^{k}_{N^{k}}
\end{equation}
\end{comment}
\begin{equation}
M = \bigcup_{k=1}^{K}\{t^{k}_{st}, \dots, t^{k}_{end}\};\ t^{k}_{1} \leq t_{st} \leq t_{end} \leq t^{k}_{N^{k}}
\end{equation}
where $t^{k}_{st}$ and $t^{k}_{end}$ denote the starting and ending timesteps, respectively, of the selected segment from schedule $k$. The overall process is explained in pseudo-code format in Alg. \ref{alg:ADSC}.
\setlength{\intextsep}{5pt}
\setlength{\textfloatsep}{5pt}
\begin{algorithm}[!h]
\caption{Control Process of the ADSC Controller}
\begin{scriptsize}
\begin{algorithmic}[1]
\State \textbf{Input:} Control space $C = \{S^{1}, S^{2}, ..., S^{K}\}$
\State \textbf{Output:} Mixed step schedule $m$
\State $m \leftarrow \emptyset, \quad k \leftarrow 1, \quad i \leftarrow 1$  \Comment{Start with the longest schedule}
\While{$t^{k}_{i}$ < $t^{k}_{N^k}$}
\If{convergence detected}
\State $k \leftarrow k + 1$ \Comment{Transition to the next schedule}
\State $i \leftarrow \arg\min\limits_{j'} |t^{k}_{j'} - t^{k-1}_{i}|$
\Else
\State $m \leftarrow m \cup \{t^{k}_{i}\}$ \Comment{Add the timestep to the mixed schedule}
\State $i \leftarrow i + 1$
\EndIf
%\If{$i \leq N_k$}
\EndWhile
\end{algorithmic}
\end{scriptsize}
\label{alg:ADSC}
\end{algorithm}
\vspace{-10pt}
\subsection{Convergence criterion}
\begin{figure}[!ht]
    \centering
    \includegraphics[width=0.99\linewidth]{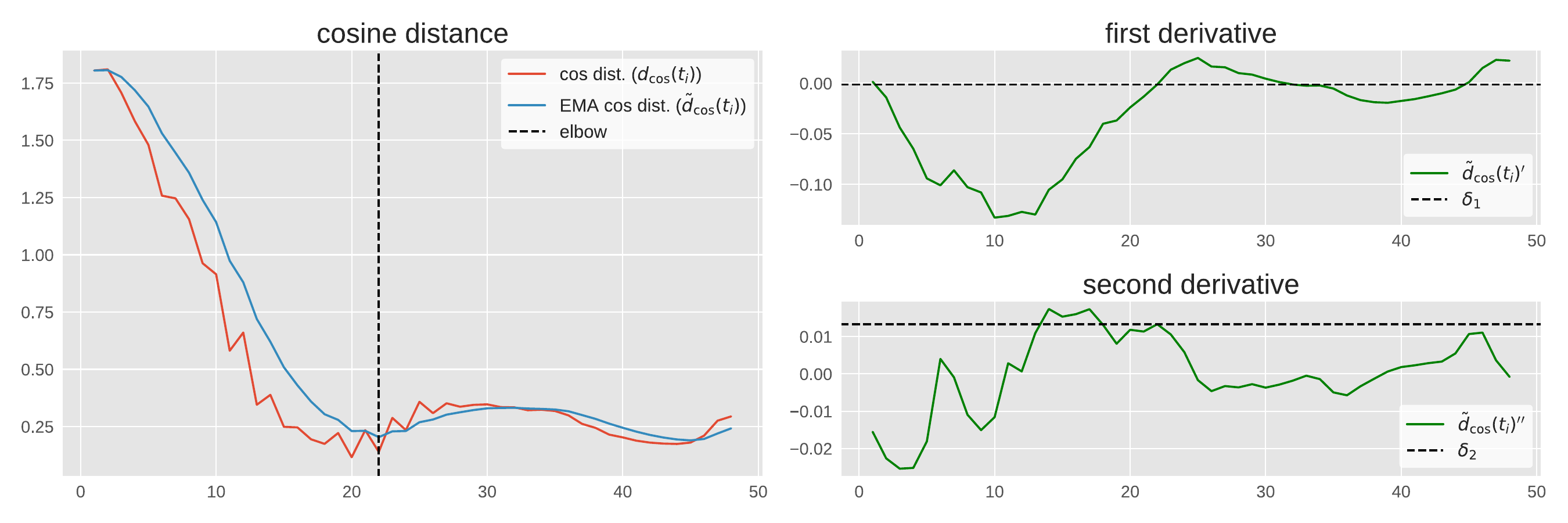}
    \vspace{-5pt}
    \caption{Plot of cosine distance between total and unconditional noise estimates across diffusion steps, along with first and second derivative curves, for a sample from the DiffusionDB dataset. The X-axis represents the diffusion steps.}
    \label{fig:cos_dist}
\end{figure}

\vspace{-5pt}
In this work we consider text-to-image generation using classifier-free guidance \cite{ho2022classifier}.
Classifier-Free Guidance generates both conditional and unconditional noise estimates (denoted by $\epsilon_{c}$ and $\epsilon_{u}$) at each denoising step, steering model outputs to match specific conditions. The conditional estimate targets the given condition (e.g., a text description), while the unconditional estimate is produced without any guiding condition. The noisy image $x_{t_{i}}$ is updated as
\begin{equation}
x_{t_{i+1}} = f(x_{t_{i}}, \underbrace{\epsilon_{u} + \gamma(\epsilon_{c} - \epsilon_{u})}_{\text{estimated noise }(\epsilon)})
\end{equation}
where $f$ is the function used to update the noisy image, and $\epsilon$ is the total noise estimate. $\gamma$ is a coefficient called ``guidance scale'', that adjusts the contribution of $\epsilon_{c}$ and $\epsilon_{u}$ while updating the image. To determine the convergence of the diffusion process, we track the cosine distance between the unconditional noise estimate and the total estimated noise $\epsilon$, which is influenced by the conditional estimate $\epsilon_{c}$. The cosine distance at each time step $t$ is denoted as $d^{(t)}{\text{cos}}(\epsilon_{u},\epsilon) = d_{\text{cos}}{(t)}$. As the diffusion progresses, $d_{\text{cos}}{(t)}$ exhibits a decline that resembles an inverse exponential function, as exemplified in Figure \ref{fig:cos_dist} for a sample drawn from DiffusionDB dataset. 
\par
The elbow point on this curve is the point after which $d_{\text{cos}}{(t)}$ does not change significantly, indicating that the denoised image has sufficiently converged to the conditional, i.e. the text prompt. At this stage, the quality of the denoised image becomes less sensitive to estimation errors resulting from large step sizes, which allows us to safely transition to the next schedule in the control space. The cosine distance curve $D_{\text{cos}}(t) = \{d_{\text{cos}}{(t_{1})},d_{\text{cos}}{(t_{2})},\ldots,d_{\text{cos}}{(t_{T})}\}$ is subject to fluctuations, and approaches that make use of the derivative values to locate the elbow can erroneously detect local minima instead.
To address this issue, we first smoothen the curve by applying exponential averaging. Let $\tilde{d}_{\text{cos}}{(t_{i})}$ denote the smoothed cosine distance at timestep $t_{i}$, which can be computed using exponential averaging:
\begin{equation}
\tilde{d}_{\text{cos}}{(t_{i})} = \alpha d_{\text{cos}}{(t_{i})} + (1 - \alpha) \tilde{d}_{\text{cos}}{(t_{i-1})}
\end{equation}
where $\alpha$ is the smoothing factor, which determines the degree of smoothing applied to the curve. Next, we examine the first and second derivatives of the smoothed cosine distance curve, denoted by $\tilde{d}_{cos}{(t_{i})}'$ and $\tilde{d}_{cos}{(t_{i})}''$ respectively:
\begin{align}
\tilde{d}_{cos}{(t_{i})}' &=  \tilde{d}_{cos}{(t_{i})} - \tilde{d}_{cos}{(t_{i-1})} \\
\tilde{d}_{cos}{(t_{i})}'' &= \tilde{d}_{cos}{(t_{i+1})} + \tilde{d}_{cos}{(t_{i-1})} - 2 \times \tilde{d}_{cos}{(t_{i})}
\end{align}
If the absolute values of both derivatives fall below predefined thresholds, we signal that the elbow point has been reached.
Finally, we define the convergence criterion in terms of the first and second derivatives:
\begin{comment}

\begin{equation}
\text{\boldsymbol{C}} =
\begin{cases}
1, & \text{if}\ | \tilde{d}_{cos}{(t_{i})}' | < \delta_1 \text{ and } | \tilde{d}_{cos}{(t_{i})}'' | < \delta_2 \\
0, & \text{otherwise}
\end{cases}
\label{eq:convergence}
\end{equation}
\end{comment}
\begin{equation}
\boldsymbol{C} =
\begin{cases}
1, & \text{if } |\tilde{d}_{\cos}(t_{i})'| < \delta_1 \text{ and } |\tilde{d}_{\cos}(t_{i})''| < \delta_2, \\
0, & \text{otherwise}.
\end{cases}
\label{eq:convergence}
\end{equation}

As expressed in Eq. \ref{eq:convergence} the convergence criterion is set to 1 (indicating convergence) if the absolute values of both the first derivative, and the second derivative,  fall below their respective threshold values, $\delta_1$ and $\delta_2$. 

\section{Experiments}
%To assess the effectiveness of our approach in enhancing the resource efficiency of text-to-image generation, 
We integrate ADSC into the Stable Diffusion pipeline \cite{rombach2022high} and compare the quality of the generated images with those produced using a fixed number of step size.
Our experiments involve utilizing PNDM \cite{liu2022pseudo} and DDIM \cite{song2020denoising} schedulers, while attaching our ADSC to these schedulers to control the generation process based on text prompts. 
In our experiments we chose the number of schedules in our control space, $K$, as 4, with each schedule containing half the number of steps as the previous one. 
This way we ensure one of every two time steps to be coinciding among consecutive schedules, allowing flexible transitions. For the guidance scale $\gamma$, we use the default value used in Stable Diffusion, that is 7.5. We evaluate the performance of our approach on two text-to-image generation datasets, namely DiffusionDB \cite{wang2022diffusiondb} and COCO \cite{lin2014microsoft}.
%DiffusionDB is a large-scale text-to-image generation dataset that comprises prompts created by real users and the corresponding images generated by the Stable Diffusion pipeline based on these prompts. These prompts describe a diverse range of realistic and abstract objects and scenes. For our evaluation, we utilize the ``2m\_random\_10k'' subset of this dataset, which consists of 10k prompt-image pairs. 
To correct spelling errors, we pre-process DiffusionDB following the procedure outlined in Wu et al. \cite{wu2023human}. %In contrast to DiffusionDB, COCO is an object recognition dataset that contains real images along with their associated object annotations. For our evaluation, we use 10k images from the COCO captions validation subset, each accompanied by a text caption describing the image content. 
%By comparing our approach's performance on these two datasets, we aim to demonstrate its effectiveness in improving the resource efficiency of text-to-image generation.
For evaulation metrics, we use the CLIP-T variant of the CLIP score \cite{hessel2021clipscore,radford2021learning}, that measures the alignment between the text prompt describing the image and the generated image itself. Later, to also assess the visual quality, we used the CLIP-I \cite{hessel2021clipscore, ruiz2023dreambooth} and DINO \cite{caron2021emerging} scores which compares the CLIP and DINO image embeddings of the generated samples with the reference images. Here we assume that images produced with many diffusion steps - mostly 50 steps - are near-optimal and use them as reference points.
%CLIP score\cite{hessel2021clipscore,radford2021learning}, also called CLIP-T  measures the cosine similarity between text features from a conditional prompt and image features of the generated image. 
%Human Preference Score (HPS) \cite{wu2023human} employs a scoring model trained to estimate human preferences for images. 
%Recognizing the limitations of these reference-free metrics, as utilised in \cite{ruiz2023dreambooth}, we incorporated reference-based metrics in our evaluation. These metrics compare generated images to reference images, assumed to be optimal representations of the text prompts. Since ground truth images are often unavailable in text-to-image generation tasks, we assume that images produced with many diffusion steps are near-optimal and use them as reference points. This ensures high-scoring images are comparable in quality to those generated with full-scale diffusion. Metrics such as CLIP-I \cite{hessel2021clipscore, ruiz2023dreambooth} and DINO \cite{caron2021emerging}, which measure feature-space similarity to reference images, are suitable for our evaluation. CLIP-I leverages CLIP image features, while DINO uses features from the ViT-S/16 DINO model.
 
\subsection{COCO results}
\begin{figure}[!ht]
  \centering
  \begin{subfigure}[b]{.45\textwidth}
  \centering
    \includegraphics[width=\textwidth]{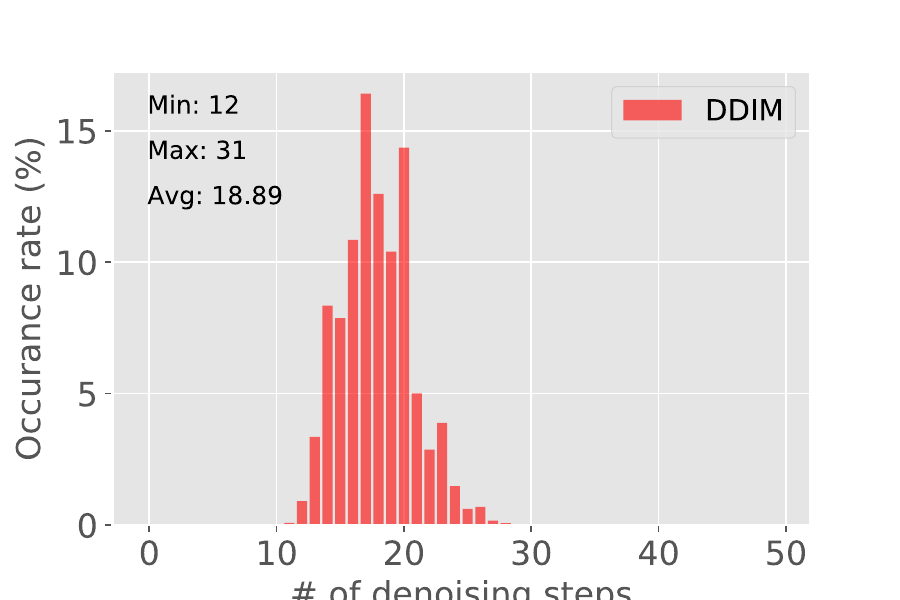}
    \caption{DDIM}
    \label{fig:eval_metrics_clip}
    \end{subfigure}
    \hspace{10pt}
    \begin{subfigure}[b]{.45\textwidth}
    \centering
    \includegraphics[width=\textwidth]{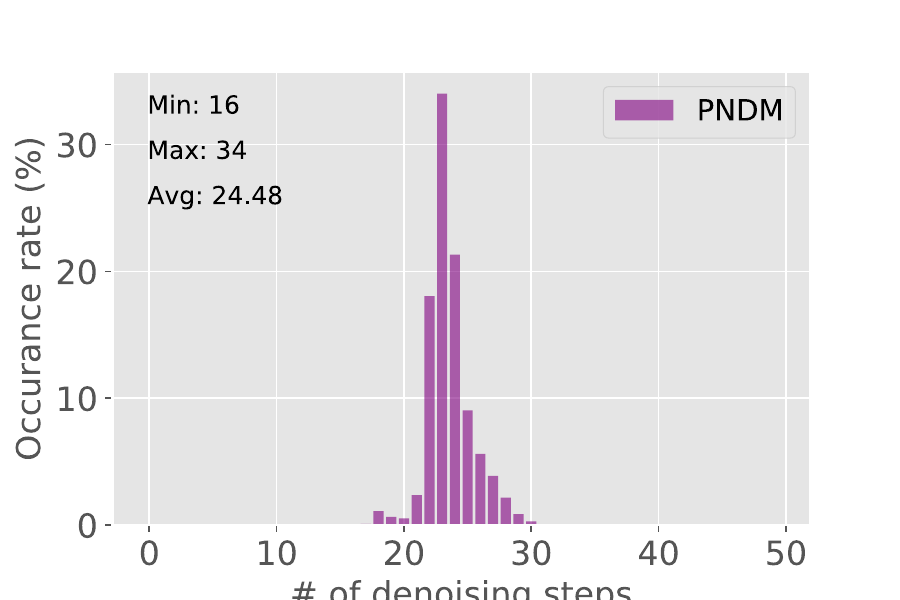}
    \caption{PNDM}
    \end{subfigure}
    \caption{Distribution of steps upon controlling DDIM and PNDM schedulers using our ADSC in COCO dataset.}
    \label{fig:COCO_dist}
   
\end{figure}

\vspace{-5pt}
\label{sec:COCO}
%\paragraph{Step distributions:} 
For our evaluation on the COCO dataset, we first processed the prompts with our ADSC attached to the PNDM \cite{liu2022pseudo} and DDIM  \cite{song2020denoising} schedulers within the Stable Diffusion v1-5 pipeline.
In contrast to these fixed-step schedulers, our method tailors the number of steps specifically for each unique image sampling task. 
Therefore, to ensure a fair comparison with fixed-step schedulers, we determine their step counts based on the minimum, maximum, and average number of steps utilized by our adaptive controller.
These values are displayed along with the visualization of the step distributions in Figure \ref{fig:COCO_dist}. 
Results indicate that the distribution of steps for both schedulers exhibits similarities to the Gaussian distribution. 
Moreover, the distribution of steps differs for each schedule, and among the two, DDIM appears to require a smaller number of steps when controlled by our ADSC.
The ADSC-controlled DDIM resulted in $18.89$ steps taken on average per prompt while PNDM led to $24.48$ steps on average.
These step values are rounded and used for DDIM and PNDM schedulers as baselines. 
%\paragraph{Quantitative results:} 
After establishing the number of steps for our baselines, we proceed to compare our method using CLIP-I, CLIP-T, and DINO scores. 
For calculating reference-based scores, namely CLIP-I and DINO, we employ images generated by fixed 50-step schedules as 50 is the default number of step used in Stable Diffusion pipeline \cite{huggingface}.   
As presented in Table \ref{tab:COCO}, the results demonstrate that, on average, ADSC attains enhanced visual quality retention for the same number of reverse diffusion steps. 
When comparing the ADSC-controlled DDIM scheduler to its 19-step fixed counterpart, we observe a $3.3\%$ ($95.70$ vs $92.40$) increase in CLIP-I score and an $8.30\%$ ($92.42$ vs $84.12$) improvement in DINO score, with an average step count of $18.89$ steps. 
Likewise, the ADSC-controlled PNDM scheduler outperforms its fixed-step counterpart, with a CLIP-I score improvement of $1.50\%$ ($94.10$ vs $92.60$) and a DINO score increase of $4.28\%$ ($89.24$ vs $84.96$). 
Notably, the ADSC-controlled DDIM also surpasses the fixed 31-step schedule. 
This can be attributed to the heterogeneous schedules offered by our adaptive controller, which are not directly comparable to homogeneous schedules containing the same number of steps.
In the case of CLIP-T, the scores remain relatively consistent across various baselines and our method, primarily because it evaluates text-image alignment.

\begin{table}[!ht]
\centering
\resizebox{1\textwidth}{!}{
\begin{tabular}{lcccc|cccc}
\hline
                                                                                                           & \multicolumn{4}{c|}{DDIM}                                                                                                                                                                                                                                                                                                      & \multicolumn{4}{c}{PNDM}                                                                                                                                                                                                                                                                                                       \\ \hline
\multicolumn{1}{l|}{Strategy}                                                                              & \cellcolor[HTML]{EFEFEF}\textbf{\begin{tabular}[c]{@{}c@{}}ADSC - ours*\\ (18.89 steps)\end{tabular}} & \begin{tabular}[c]{@{}c@{}}fixed\\ (12 steps)\end{tabular} & \cellcolor[HTML]{EFEFEF}\textbf{\begin{tabular}[c]{@{}c@{}}fixed*\\ (19 steps)\end{tabular}} & \begin{tabular}[c]{@{}c@{}}fixed\\ (31 steps)\end{tabular} & \cellcolor[HTML]{EFEFEF}\textbf{\begin{tabular}[c]{@{}c@{}}ADSC - ours*\\ (24.48 steps)\end{tabular}} & \begin{tabular}[c]{@{}c@{}}fixed\\ (16 steps)\end{tabular} & \cellcolor[HTML]{EFEFEF}\textbf{\begin{tabular}[c]{@{}c@{}}fixed*\\ (24 steps)\end{tabular}} & \begin{tabular}[c]{@{}c@{}}fixed\\ (34 steps)\end{tabular} \\ \hline
\multicolumn{1}{l|}{CLIP-I ($\uparrow$)}                                                                   & \cellcolor[HTML]{EFEFEF}\textbf{95.70}                                                                & 90.00                                                      & \cellcolor[HTML]{EFEFEF}92.40                                                                & 95.20                                                      & \cellcolor[HTML]{EFEFEF}\textbf{94.10}                                                                & 91.30                                                      & \cellcolor[HTML]{EFEFEF}92.60                                                                & 94.50                                                      \\ \hline
\multicolumn{1}{l|}{CLIP-T ($\uparrow$)}                                                                   & \cellcolor[HTML]{EFEFEF}\textbf{31.40}                                                                & 31.40                                                      & \cellcolor[HTML]{EFEFEF}\textbf{31.40}                                                       & 31.40                                                      & \cellcolor[HTML]{EFEFEF}\textbf{31.40}                                                                & 31.30                                                      & \cellcolor[HTML]{EFEFEF}\textbf{31.40}                                                       & 31.30                                                      \\ \hline
\multicolumn{1}{l|}{DINO ($\uparrow$)}                                                                     & \cellcolor[HTML]{EFEFEF}\textbf{92.42}                                                                & 78.78                                                      & \cellcolor[HTML]{EFEFEF}84.12                                                                & 90.72                                                      & \cellcolor[HTML]{EFEFEF}\textbf{89.24}                                                                & 81.06                                                      & \cellcolor[HTML]{EFEFEF}84.96                                                                & 89.20                                                      \\ \hline
\multicolumn{1}{l|}{\begin{tabular}[c]{@{}l@{}}Avg wallclock \\ time (sec/img)($\downarrow$)\end{tabular}} & \cellcolor[HTML]{EFEFEF}0.79                                                                          & 0.55                                                       & \cellcolor[HTML]{EFEFEF}0.78                                                                 & 1.23                                                       & \cellcolor[HTML]{EFEFEF}1.06                                                                          & 0.71                                                       & \cellcolor[HTML]{EFEFEF}1.06                                                                 & 1.40                                                       \\ \hline
\end{tabular}
}
\caption{Comparison between ADSC and baseline fixed-step schedulers using various metrics on the COCO dataset. (*) highlights our method and the main baselines.}
\label{tab:COCO}
\end{table}
\par
%Our approach introduces additional computation to each step by calculating cosine similarities and their derivatives; therefore reporting only the average step reduction is insufficient for a comprehensive evaluation. 

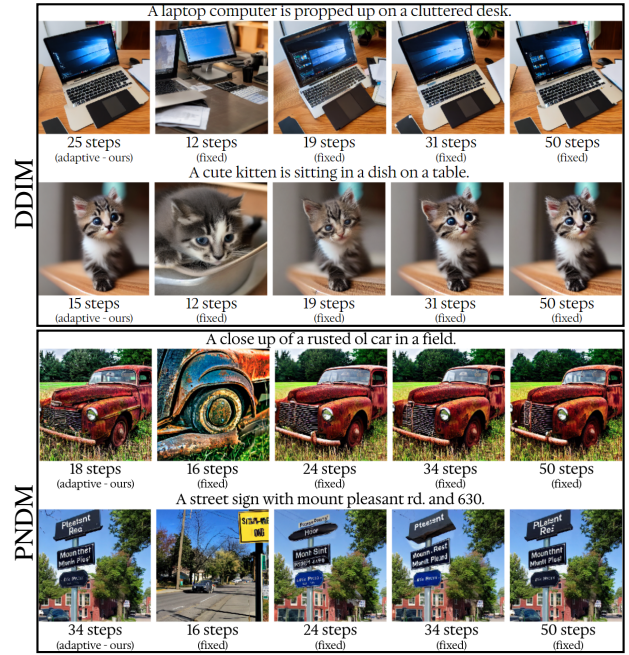
\begin{figure}[!b]  
    \vspace{10pt}
    \centering  
    \begin{tikzpicture}  
        \node[inner sep=0pt] (ddim1) {  
            \begin{subfigure}{0.9\textwidth}  
                \centering  
                \includegraphics[width=\textwidth]{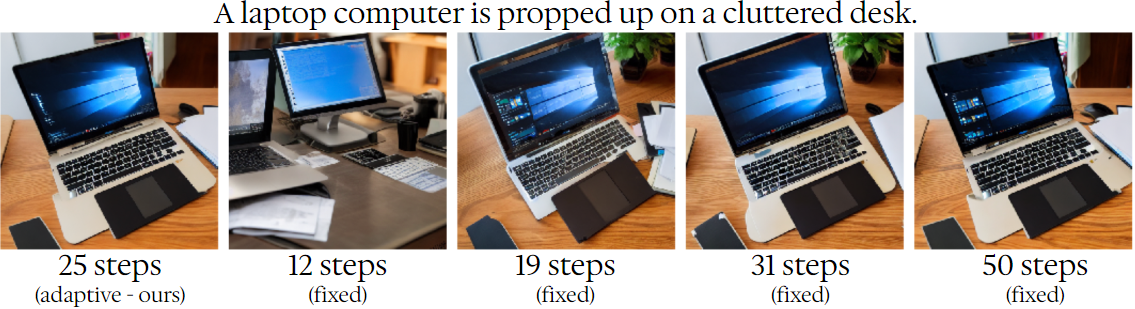}  
            \end{subfigure}  
        };  
        \node[inner sep=0pt, below=0pt of ddim1] (ddim2) {  
            \begin{subfigure}{0.9\textwidth}  
                \centering  
                \includegraphics[width=\textwidth]{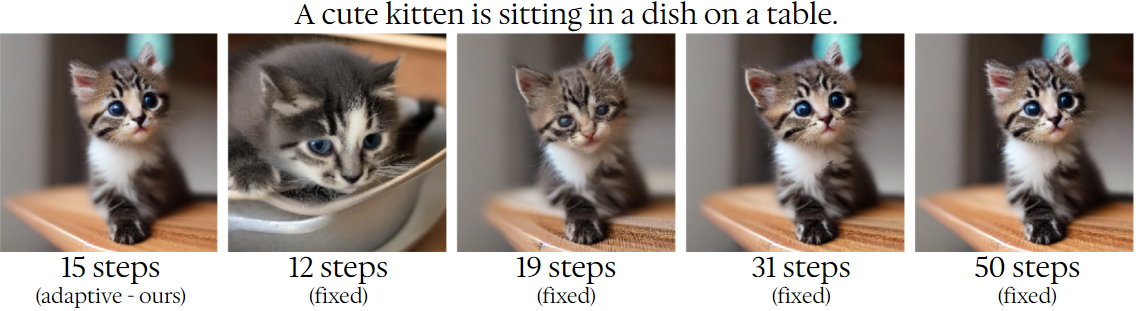}  
            \end{subfigure}  
        };  
        \draw[black, thick] (ddim1.north west) rectangle (ddim2.south east);  
        \node[rotate=90, anchor=east, yshift=5pt] at ($(ddim1.west)!0.4!(ddim2.west)$) {DDIM};  
    \end{tikzpicture}  
  
    \begin{tikzpicture}  
        \node[inner sep=0pt] (pndm1) {  
            \begin{subfigure}{0.9\textwidth}  
                \centering  
                \includegraphics[width=\textwidth]{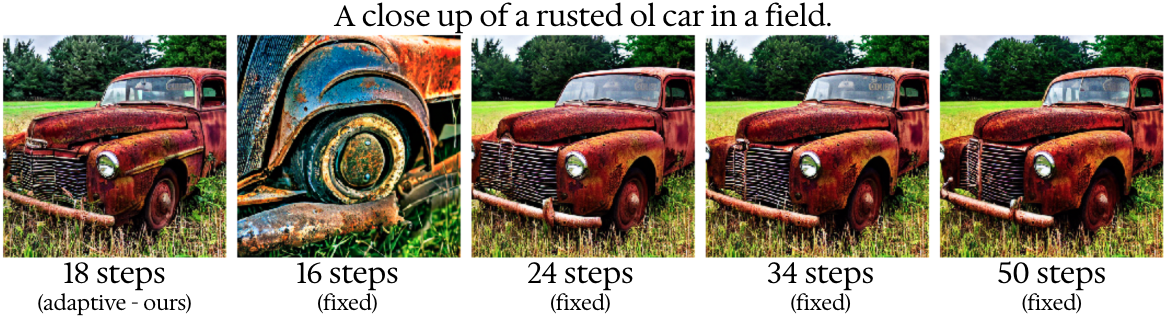}  
            \end{subfigure}  
        };  
        \node[inner sep=0pt, below=0pt of pndm1] (pndm2) {  
            \begin{subfigure}{0.9\textwidth}  
                \centering  
                \includegraphics[width=\textwidth]{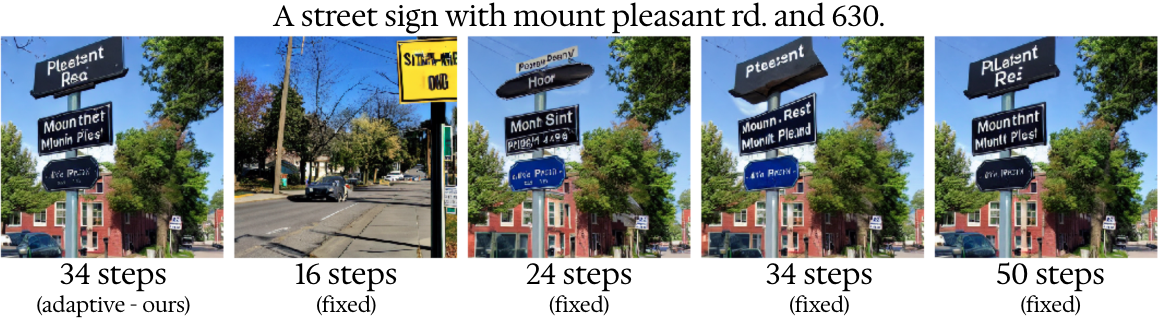}  
            \end{subfigure}  
        };  
        \draw[black, thick] (pndm1.north west) rectangle (pndm2.south east);  
        \node[rotate=90, anchor=east, yshift=5pt] at ($(pndm1.west)!0.4!(pndm2.west)$) {PNDM};  
    \end{tikzpicture}  
      
    \caption{Text prompts from COCO and images generated by our ADSC, and schedulers fixed number of steps.}
    \label{fig:coco_imgs}  
\end{figure}
To ensure a comprehensive assessment, we also measure the average wall-clock times (latencies) per generated image. The results reveal that the latency of ADSC is nearly identical to that of fixed schedulers, with only minor differences up to two decimal places. For instance, the ADSC-controlled DDIM scheduler took an average of 0.79 seconds for $18.89$ steps, which is equal to the runtime of the 19-step DDIM schedule. 
%This outcome is anticipated, as the computation of first and second derivatives for discrete signals primarily involves basic addition and subtraction operations. The running average computation additionally requires scalar multiplications, which are also relatively simple operations. Although computing cosine distances is more complex compared to other calculations, as it involves the dot product of two vectors, the overall latency is not significantly affected due to the relatively small dimensions of image latents.
%\paragraph{Qualitative results:}
Lastly, we provide example images generated by our ADSC and fixed baselines in Figure \ref{fig:coco_imgs}. 
The first and last two rows are obtained by DDIM and PNDM schedulers, respectively.
An initial observation reveals that, in general, ADSC resulted in higher-quality images within a smaller number of timesteps compared to using a fixed step scheduler. Moreover, the prompts given in the first and last rows required more steps to generate images of sufficient quality compared to others. This can be attributed to the fact that the objects described in these prompts inherently contain intricate details, such as the keyboard and home screen of a laptop or street names written on a sign. 
Conversely, cats, cars, and tables are relatively more plain objects. 
The commonality of the objects and their presence in the training dataset of the diffusion model can be another contributing factor. 
%Another observation is that in some cases even the 50-step generated reference images contain defects. However, as our goal is not to improve the diffusion model itself, we consider this image as optimal and assess other images in relation to it. 
\subsection{DiffusionDB results}
\begin{figure}[!ht]
  \centering
  \begin{subfigure}[b]{.45\textwidth}
  \centering
    \includegraphics[width=\textwidth]{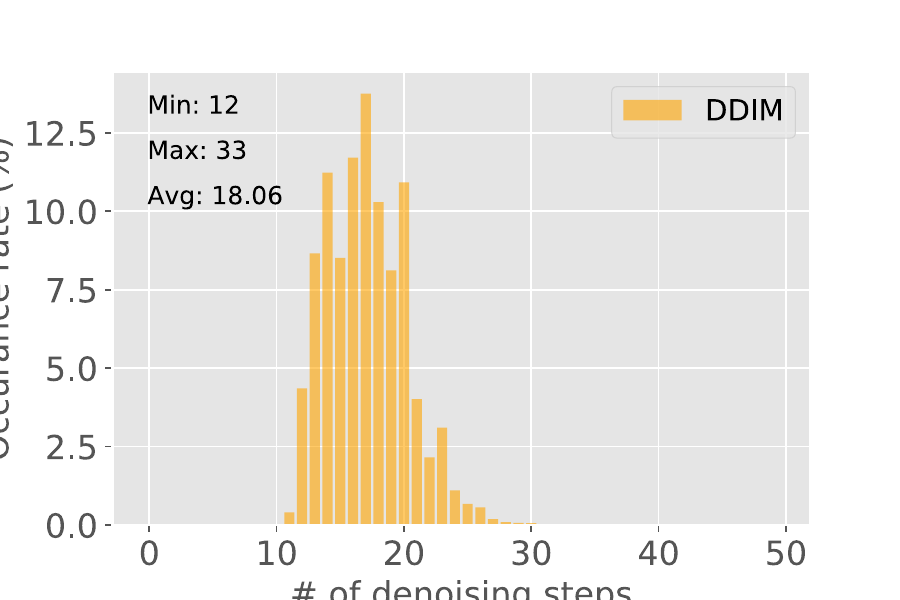}
    \caption{DDIM}
    \label{fig:eval_metrics_clip}
    \end{subfigure}
    \hspace{10pt}
    \begin{subfigure}[b]{.45\textwidth}
    \centering
    \includegraphics[width=\textwidth]{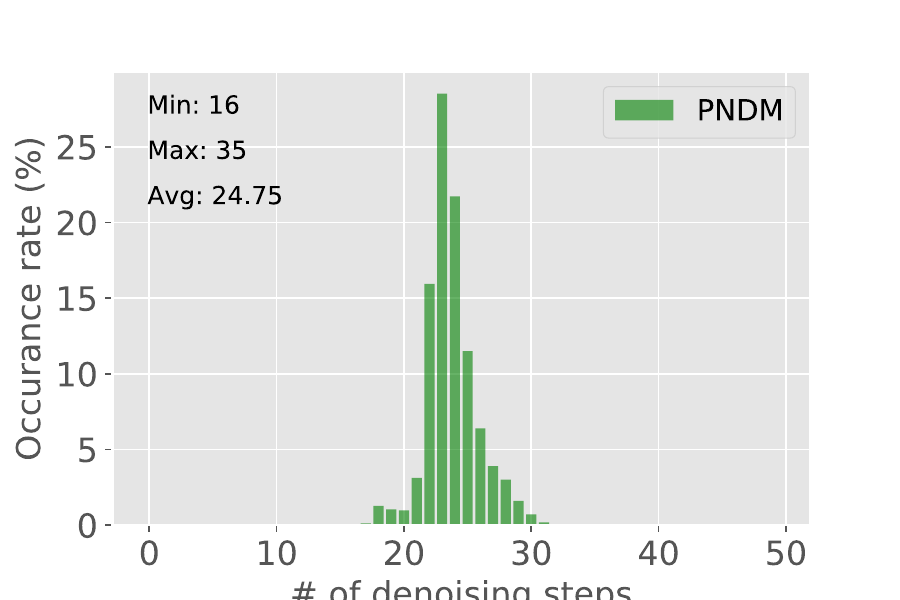}
    \caption{PNDM}
    \end{subfigure}
    \vspace{-10pt}
    \caption{Distribution of steps upon controlling DDIM and PNDM schedulers using our ADSC in DiffusionDB dataset.}
    \label{fig:diffusionDB_dist}
   
\end{figure}

We followed the same procedures described in Section \ref{sec:COCO} to evaluate our method on the DiffusionDB dataset. The distribution of steps exhibits differences compared to COCO experiments, as seen in Figure \ref{fig:diffusionDB_dist}. These differences are also reflected in the avg, min, and max statistics.
%As predicted, the distribution of steps exhibits differences compared to that observed in COCO experiments. The most noticeable difference is that, the occurance rate of number of steps taken by ADSC-controlled DDIM scheduler is more uniformly distributed as seen in Figure \ref{fig:diffusionDB_dist}. As for PNDM, the distribution again follows a Gaussian as in COCO experiments, yet the probabilities are less skewed towards certain step counts. These differences are also reflected in the avg, min, and max statistics.
Overall, both adaptive and fixed scheduling approaches saw an improvement in CLIP-T scores beyond those observed in the COCO experiments. 
Upon coupling ADSC with DDIM, we observed that the improvement margins of CLIP-I and DINO remain significant, which are $3.2\%$ ($95.10$ vs. $91.90$) and $6.4\%$ ($89.36$ vs. $82.96$) respectively. 
As for the PNDM scheduler, prompt-adaptive control did not lead to any noticeable improvement in CLIP-I and caused the DINO score to drop by $1.59\%$. 
This suggests that ADSC is more compatible with DDIM and more consistently improves performance when coupled with it. 
However, it is also worth noting that ADSC did not degrade PNDM performance significantly either. 
Therefore, considering the improvements seen in benchmarks like COCO, it remains beneficial to employ ADSC to enhance PNDM as well.
\begin{table}[!ht]
\centering
\resizebox{1\textwidth}{!}{
\begin{tabular}{lcccc|cccc}
\hline
                                                                                                           & \multicolumn{4}{c|}{DDIM}                                                                                                                                                                                                                                                                                                      & \multicolumn{4}{c}{PNDM}                                                                                                                                                                                                                                                                                                       \\ \hline
\multicolumn{1}{l|}{Strategy}                                                                              & \cellcolor[HTML]{EFEFEF}\textbf{\begin{tabular}[c]{@{}c@{}}ADSC - ours*\\ (18.06 steps)\end{tabular}} & \begin{tabular}[c]{@{}c@{}}fixed\\ (12 steps)\end{tabular} & \cellcolor[HTML]{EFEFEF}\textbf{\begin{tabular}[c]{@{}c@{}}fixed*\\ (18 steps)\end{tabular}} & \begin{tabular}[c]{@{}c@{}}fixed\\ (33 steps)\end{tabular} & \cellcolor[HTML]{EFEFEF}\textbf{\begin{tabular}[c]{@{}c@{}}ADSC - ours*\\ (24.75 steps)\end{tabular}} & \begin{tabular}[c]{@{}c@{}}fixed\\ (16 steps)\end{tabular} & \cellcolor[HTML]{EFEFEF}\textbf{\begin{tabular}[c]{@{}c@{}}fixed*\\ (25 steps)\end{tabular}} & \begin{tabular}[c]{@{}c@{}}fixed\\ (35 steps)\end{tabular} \\ \hline
\multicolumn{1}{l|}{CLIP-I ($\uparrow$)}                                                                   & \cellcolor[HTML]{EFEFEF}\textbf{95.10}                                                                & 89.40                                                      & \cellcolor[HTML]{EFEFEF}91.90                                                                & 95.20                                                      & \cellcolor[HTML]{EFEFEF}\textbf{93.70}                                                                & 89.90                                                      & \cellcolor[HTML]{EFEFEF}93.60                                                                & 95.70                                                      \\ \hline
\multicolumn{1}{l|}{CLIP-T ($\uparrow$)}                                                                   & \cellcolor[HTML]{EFEFEF}33.70                                                                         & 33.70                                                      & \cellcolor[HTML]{EFEFEF}\textbf{33.90}                                                       & 34.00                                                      & \cellcolor[HTML]{EFEFEF}\textbf{33.80}                                                                & 33.30                                                      & \cellcolor[HTML]{EFEFEF}33.70                                                                & 33.80                                                      \\ \hline
\multicolumn{1}{l|}{DINO ($\uparrow$)}                                                                     & \cellcolor[HTML]{EFEFEF}\textbf{89.36}                                                                & 77.01                                                      & \cellcolor[HTML]{EFEFEF}82.96                                                                & 90.90                                                      & \cellcolor[HTML]{EFEFEF}85.50                                                                         & 79.02                                                      & \cellcolor[HTML]{EFEFEF}\textbf{87.09}                                                       & 87.21                                                      \\ \hline
\multicolumn{1}{l|}{\begin{tabular}[c]{@{}l@{}}Avg wallclock \\ time (sec/img)($\downarrow$)\end{tabular}} & \cellcolor[HTML]{EFEFEF}0.76                                                                          & 0.55                                                       & \cellcolor[HTML]{EFEFEF}0.75                                                                 & 1.32                                                       & \cellcolor[HTML]{EFEFEF}1.06                                                                          & 0.71                                                       & \cellcolor[HTML]{EFEFEF}1.06                                                                 & 1.42                                                       \\ \hline
\end{tabular}
}
\caption{Comparison between ADSC and baseline fixed-step schedulers using various metrics on the DifusionDB dataset. (*) highlights our method and the main baselines.}
\label{tab:DiffusionDB}
\end{table}
%\paragraph{Qualitative results:}
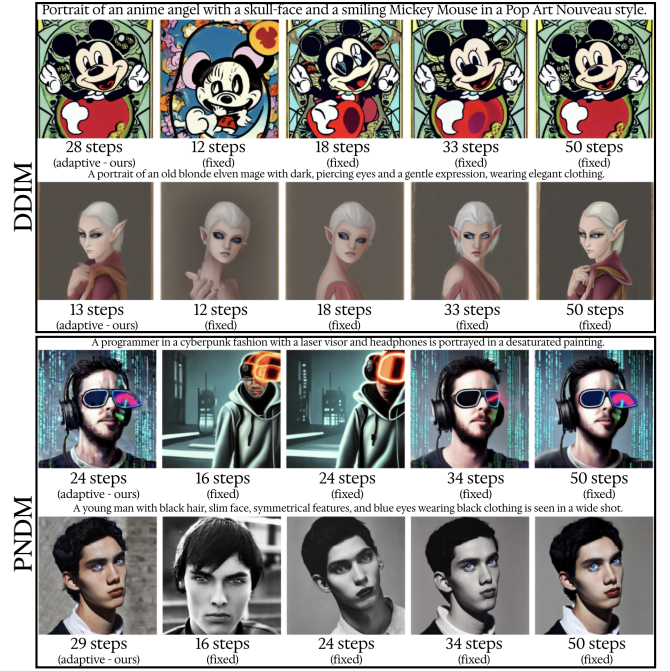
\begin{figure}[!ht]  
    \centering  
    \begin{tikzpicture}  
        \node[inner sep=0pt] (ddim1) {  
            \begin{subfigure}{0.95\textwidth}  
                \centering  
                \includegraphics[width=\textwidth]{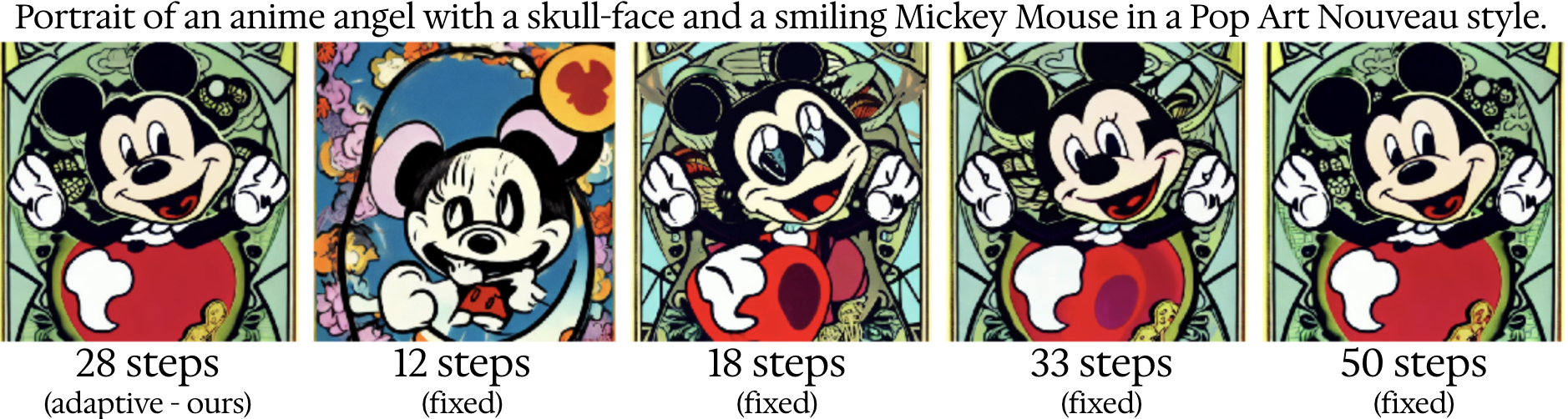}  
            \end{subfigure}  
        };  
        \node[inner sep=0pt, below=0pt of ddim1] (ddim2) {  
            \begin{subfigure}{0.95\textwidth}  
                \centering  
                \includegraphics[width=\textwidth]{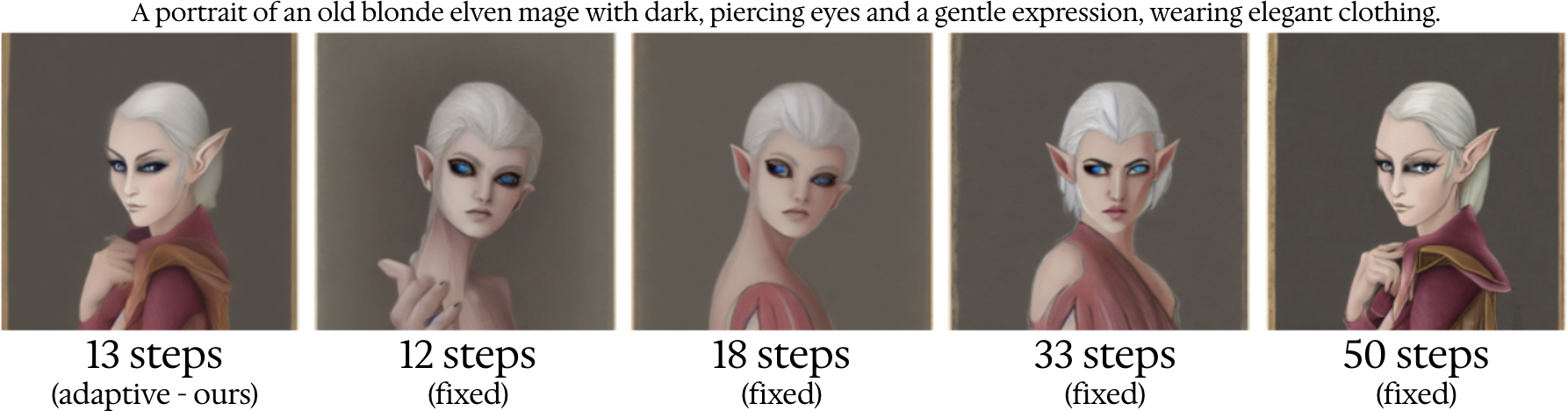}  
            \end{subfigure}  
        };  
        \draw[black, thick] (ddim1.north west) rectangle (ddim2.south east);  
        \node[rotate=90, anchor=east, yshift=5pt] at ($(ddim1.west)!0.4!(ddim2.west)$) {DDIM};  
    \end{tikzpicture}  
  
    \begin{tikzpicture}  
        \node[inner sep=0pt] (pndm1) {  
            \begin{subfigure}{0.95\textwidth}  
                \centering  
                \includegraphics[width=\textwidth]{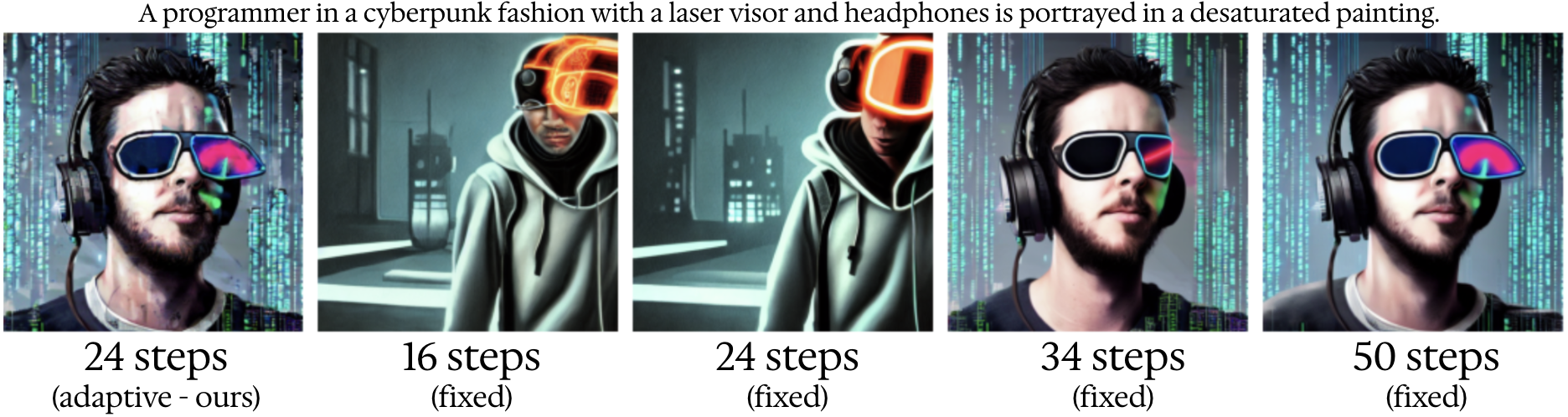}  
            \end{subfigure}  
        };  
        \node[inner sep=0pt, below=0pt of pndm1] (pndm2) {  
            \begin{subfigure}{0.95\textwidth}  
                \centering  
                \includegraphics[width=\textwidth]{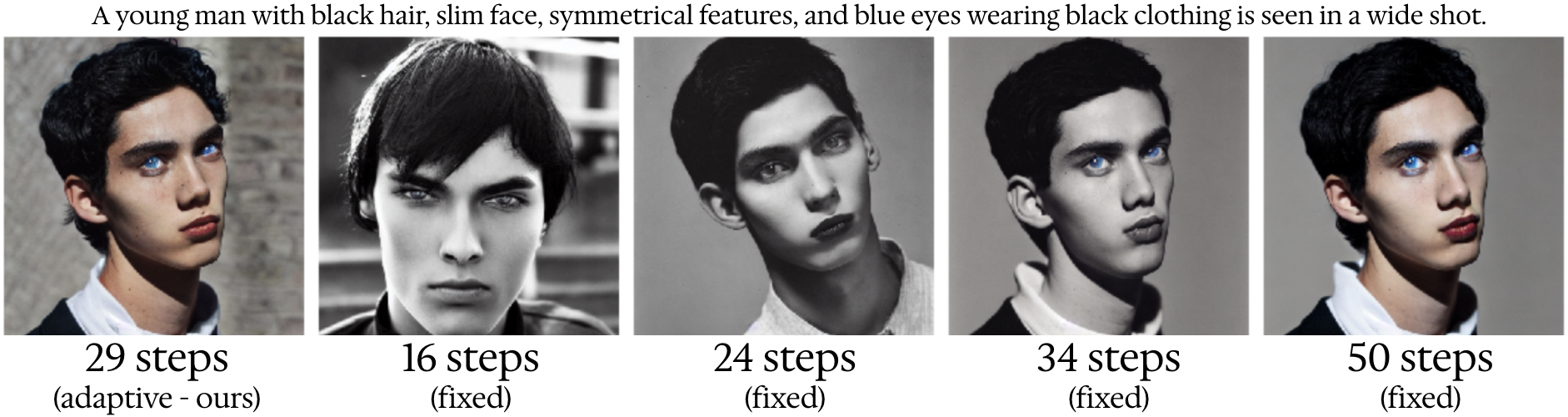}  
            \end{subfigure}  
        };  
        \draw[black, thick] (pndm1.north west) rectangle (pndm2.south east);  
        \node[rotate=90, anchor=east, yshift=5pt] at ($(pndm1.west)!0.4!(pndm2.west)$) {PNDM};  
    \end{tikzpicture}  
      
    \caption{Text prompts from DiffusionDB and images generated by our ADSC, and schedulers with fixed numbers of steps.}  
    \label{fig:diffusionDB_imgs}  
\end{figure}
Figure \ref{fig:diffusionDB_imgs} contains images generated based on DiffusionDB prompts. The first two and last two rows are generated using DDIM and PNDM, respectively. For all four prompts, ADSC produced high-quality images, while several problems can be noticed in the images generated by fixed schedules. 
The first one is the misalignment with the text prompt, which can be observed in the last two rows: The 16 and 24-step generated images from the prompt "A programmer in a cyberpunk..."  do not contain distinctive features of a programmer, and the 16, 24-step generated images from the prompt "A young man with black..." miss the color information of facial features described in the prompt as they are in grayscale. 
Although the 34-step generated image contains blue eyes, it is inconsistent with the grayscale theme. 
The second problem is the visual defects which can be seen in the first, second, and third rows.
Finally, we expand our evaluation by comparing our method against fixed step schedules from two other schedulers, DDPM \cite{ho2020denoising}  and LMS \cite{karras2022elucidating}. We employ our ADSC combined with the DDIM scheduler for adaptive scheduling, as it yielded the best results. Table \ref{tab:var_schedulers} presents the findings.
\begin{table}[!ht]
\resizebox{0.6\textwidth}{!}{
\begin{tabular}{l|ccc|ccc}
\hline
Dataset                                                      & \multicolumn{3}{c|}{\begin{tabular}[c]{@{}c@{}}DiffusionDB\\ (50 $\rightarrow$ 18 steps)\end{tabular}} & \multicolumn{3}{c}{\begin{tabular}[c]{@{}c@{}}COCO\\ (50 $\rightarrow$ 19 steps)\end{tabular}} \\ \hline
Metric                                                       & \multicolumn{1}{c|}{CLIP-I}             & \multicolumn{1}{c|}{CLIP-T}             & DINO               & \multicolumn{1}{c|}{CLIP-I}           & \multicolumn{1}{c|}{CLIP-T}          & DINO            \\ \hline
DDPM \cite{ho2020denoising}                                  & \multicolumn{1}{c|}{86.7}               & \multicolumn{1}{c|}{33.5}               & 69.68              & \multicolumn{1}{c|}{88.5}             & \multicolumn{1}{c|}{31.3}            & 74.8            \\
PNDM \cite{liu2022pseudo}                                    & \multicolumn{1}{c|}{90.8}               & \multicolumn{1}{c|}{33.5}               & 80.69              & \multicolumn{1}{c|}{91.7}             & \multicolumn{1}{c|}{31.3}            & 82.3            \\
DDIM \cite{song2020denoising}                                & \multicolumn{1}{c|}{91.9}               & \multicolumn{1}{c|}{\textbf{33.9}}      & 83.0               & \multicolumn{1}{c|}{92.4}             & \multicolumn{1}{c|}{\textbf{31.4}}   & 84.1            \\
LMS \cite{karras2022elucidating}                            & \multicolumn{1}{c|}{94.1}               & \multicolumn{1}{c|}{33.7}               & 88.2               & \multicolumn{1}{c|}{94.8}             & \multicolumn{1}{c|}{31.3}            & 90.5            \\ \hline
\begin{tabular}[c]{@{}l@{}}DDIM + ADSC\\ (ours)\end{tabular} & \multicolumn{1}{c|}{\textbf{95.1}}      & \multicolumn{1}{c|}{33.7}               & \textbf{89.4}      & \multicolumn{1}{c|}{\textbf{95.7}}    & \multicolumn{1}{c|}{\textbf{31.4}}   & \textbf{92.4}   \\ \hline
\end{tabular}}
\caption{Comparison with various schedulers}
\label{tab:var_schedulers}
\end{table}

\begin{comment}

\section{Limitations}
While our proposed ADSC demonstrates promising improvements in time efficiency and visual quality for text-to-image generation tasks, it is important to acknowledge its limitations. 
First, the convergence criterion utilized by our step scheduler controller is based on conditional guidance, in our case, text prompts describing images. As a result, our method is specifically tailored for conditional generation tasks using diffusion models. This implies that the applicability of our approach is limited to scenarios where conditional input is available, and it may not be directly applicable or effective for unconditional generation tasks. Moreover, we introduce a single heuristic criterion, the cosine distance between the conditional and unconditional update directions, to guide the control process. While this criterion has proven effective in our experiments, there is potential for exploring alternative criteria and also learning-based approaches to further improve the performance of the scheduler controller. 
\end{comment}

\section{Conclusion}
%In this paper, we introduced ADSC, designed to address the inefficiency of fixed denoising schedules in text-to-image diffusion generative models. Our method dynamically adjusts the number of denoising steps required to generate high-quality images based on the specific characteristics of each text prompt, thereby optimizing computational resources and time efficiency without compromising significant visual quality. Experiments on the COCO and DiffusionDB datasets, we have demonstrated the effectiveness of our approach in reducing the average number of timesteps needed for image generation by over $50\%$ in some cases, while still preserving the visual fidelity of the output images. Importantly, our method does not necessitate additional model training, as it dynamically adapts the denoising schedule during runtime based on the convergence of the generation process.

We introduced ADSC to address the inefficiency of fixed denoising schedules in text-to-image diffusion generative models. Our method dynamically adjusts the number of denoising steps required to generate high-quality images based on the specific characteristics of each text prompt, thereby optimizing computational resources and time efficiency without compromising significant visual quality. Experiments have demonstrated the effectiveness of our approach in reducing the average number of diffusion time steps, while still preserving the visual fidelity. Importantly, our method does not necessitate additional model training. One limitation is the need for manual tuning of hyper-parameters, such as gradient thresholds, which can be time-consuming process. Future work can explore adaptive thresholding and learned convergence criteria using reinforcement learning methods.

\vfill\pagebreak

\bibliographystyle{IEEEbib}
\bibliography{refs}

\end{document}